# Improving Predictive Confidence in Medical Imaging via Online Label Smoothing

*Kushan* Choudhury[1][1], *Shubhrodeep* Roy[1], *Ankur* Chanda[1], *Shubhajit* Biswas[1] and *Somenath* Kuiry[1]

[1]Department of CSE(AIML), Institute of Engineering & Management, Kolkata, India

**Abstract.** Deep learning models, especially convolutional neural networks, have achieved impressive results in medical image classification. However, these models often produce overconfident predictions, which can undermine their reliability in critical healthcare settings. While traditional label smoothing offers a simple way to reduce such overconfidence, it fails to consider relationships between classes by treating all non-target classes equally. In this study, we explore the use of Online Label Smoothing (OLS) , a dynamic approach that adjusts soft labels throughout training based on the model's own prediction patterns. We evaluate OLS on the large-scale RadImageNet dataset using three widely-used architectures: ResNet-50, MobileNetV2, and VGG-19. Our results show that OLS consistently improves both Top-1 and Top-5 classification accuracy compared to standard training methods, including hard labels, conventional label smoothing, and teacher-free knowledge distillation. In addition to accuracy gains, OLS leads to more compact and well-separated feature embeddings, indicating improved representation learning. These findings suggest that OLS not only strengthens predictive performance but also enhances calibration, making it a practical and effective solution for developing trustworthy AI systems in the medical imaging domain.

## 1      Introduction

Deep neural networks (DNNs) deliver state-of-the-art performance on most visual tasks but are commonly overconfident in their probability estimates, hence poorly calibrated. Miscalibration of this nature can compromise clinical trust and decision‑making in safety-critical applications such as medical imaging. For instance, recent convolutional networks have been found to become drastically overconfident on training examples [1]. To counter domain-specific problems, large medical image datasets have been created. Significantly, RadImageNet is a corpus of 1.35 million annotated radiology images (CT, MRI, and ultrasound) for musculoskeletal, neurologic, oncologic, abdominal, endocrine, and pulmonary pathologies [2]. Pretrained models with RadImageNet perform better on various radiologic tasks than ImageNet models [2], reflecting the appropriateness of the dataset to develop trustworthy medical classifiers. Large-scale, multi-modal data as such present the perfect benchmark for calibrating medical AI methods.

---

[1] Corresponding author: kushanchoudhury2003@gmail.com

One popular method to counteract overconfidence is label smoothing (LS). LS substitutes the hard one-hot labels with a weighted average of the ground-truth label and a uniform distribution across all classes. This simple alteration serves as a regularizer, discouraging the network from outputting peak probability over one class [1]. Empirically, LS has been demonstrated to decrease overfitting and alleviate extreme confidence by "softening" the targets. But standard LS has one significant shortcoming: it gives the same low probability to all non-target classes [1]. That is, LS disregards semantic or structural connections between classes. For example, a cat image should presumably be viewed as more similar to a dog than to a car, but uniform LS would assign probability equally to both [1]. This uniform smoothing can reduce LS's performance when class similarities are strongly non-uniform, as they so commonly are in radiology (e.g., anatomical differences among related organs vs. unrelated disease).

To overcome these limitations, a number of adaptive smoothing methods have been introduced. For instance, [3]. introduce confusion‑penalty label smoothing [3]. This technique monitors the model's validation confusion matrix each epoch and redistributes the smoothing mass preferably to the most confused-with classes [3]. Non-target classes that have semantic or visual similarity to the ground truth class (and hence are often confused) thus get greater soft-label weight. In another direction, [4] introduce GeoLS (Geodesic Label Smoothing) for image segmentation. GeoLS uses the geodesic distance transform of the image to modify labels: it redistributes label probabilities in a smooth manner across pixel boundaries based on image gradients [4]. Through the use of spatial context, GeoLS more effectively captures loose boundaries between classes, a key factor in dense medical prediction tasks [4]. They reflect the trend towards data-informed smoothing: instead of an task-independent uniform prior, the smoothing distribution is informed by the task (confusion patterns, image geometry, etc.).

Based on this concept of dynamic labeling, Online Label Smoothing (OLS) has been proposed to generate the smoothing distribution during training time. [1] define OLS as an on-the-fly algorithm that maintains a moving soft-label distribution for each class based on the network's own predictions [1]. Specifically, for every epoch, OLS calculates the empirical distribution of the model's outputs for every true class and adopts this moving distribution (instead of a constant uniform vector) as the target soft-label [1]. This results in more reasonable supervision: classes that the model is likely to mix up with the actual class receive more probability mass in the label, and dissimilar classes remain with approximately zero weight. [1] demonstrated that OLS can realize huge improvements: on the benchmarks of CIFAR-100 and ImageNet it improved classification accuracy, and more significantly it made models significantly more robust to corrupted labels than vanilla LS [1]. These findings suggest that OLS's data-driven smoothing is capable of better adapting predictions by reconciling training targets with observed ambiguity.

In this paper, we apply OLS to the radiological imaging area through the RadImageNet dataset [2]. RadImageNet's 1.35 million images cover CT, MRI, and ultrasound of various body regions, and pretrained RadImageNet models have made substantial AUC gains over ImageNet models in applications like thyroid nodule detection, meniscus tear classification, and pulmonary disease screening [2]. RadImageNet is a realistic site to evaluate calibration methods due to its size and diversity since it can be used to train deep classifiers from scratch in the medical field, in contrast to smaller annotated datasets. We will train convolutional classifiers on RadImageNet both with and without OLS (and compare to standard LS), and test predictive calibration (e.g., via reliability diagrams and expected calibration error) as well as overall accuracy.

Overall, this paper explores whether Online Label Smoothing can enhance the calibration and reliability of medical image classifiers. Previous research has indicated that static label smoothing enhances model calibration but that dynamic strategies such as OLS are yet to be evaluated in radiology. We expect that by evolving labels based on the model's confusion, OLS will generate probability estimates that are more calibrated toward actual uncertainty. We seek to establish whether models trained with OLS are more predictive, confidence-calibrated and trustworthy on medical image tasks compared to traditionally trained models.

## 2      Related Works

Deep neural networks often achieve strong accuracy yet remain miscalibrated, a problem amplified in clinical pipelines with class imbalance, limited data, and imperfect labels; studies spanning computer vision and lung-cancer CT show that methods effective in natural images may transfer poorly to medical imaging, with some combinations even pushing models toward under-confidence, underscoring the need for domain-aware calibration strategies [8]. A common training-time regularizer is label smoothing (LS), which mixes one-hot targets with a uniform prior and typically reduces overconfidence while improving generalization; nonetheless, the uniform allocation ignores structure among confusable classes and can blunt information carried by non-target logits [1]. Aside from accuracy, LS was demonstrated to implicitly enhance calibration but can degrade knowledge distillation through condensing penultimate-layer clusters and removing inter-class logit design, making clear when and why LS assists or injures downstream transfer [6]. In medical imaging, data-aware variants replace uniform non-target mass with informed priors: confusion-penalty LS (CPLS) shifts probability toward empirically confusable classes using a running confusion matrix (motivated by diagnostic ambiguity in histopathology) [3], and LS+ (informed LS) improves calibration/retention behavior over standard LS on dermoscopy and chest-x-ray cohorts by shaping confidence on incorrect predictions [7]; for dense prediction, GeoLS injects spatial context via geodesic distance maps to form intensity- and boundary-aware soft labels, yielding consistent gains on brain tumor, abdominal organ, and prostate segmentation benchmarks [4]. Under label noise, LS behaves like shrinkage and can compete with loss-correction methods, offering a pragmatic choice when annotations are coarse or discordant—conditions typical in healthcare datasets [5]. Against this backdrop, Online Label Smoothing (OLS) replaces the uniform non-target mass with class-conditioned soft targets that evolve during training, estimated from a model's own prediction statistics; by encoding empirical confusability, OLS improves classification on CIFAR-100 and enhances robustness to noisy labels—properties that align well with medical imaging, where similar pathologies and noisy supervision are routine [1].

## 3      Research Methodology

We implemented Online Label Smoothing (OLS) for medical image classification using the RadImageNet dataset, which contains 1.35 million annotated CT, MRI, and ultrasound images across diverse anatomical and pathological categories. Our pipeline was designed to adaptively update label distributions during training and evaluate whether this dynamic supervision improves predictive calibration and accuracy.

Data and preprocessing. Images mentioned in list files were read using OpenCV, resized to 224×224 pixels, normalized to float, and scaled linearly to the range [−1, 1]. For robustness, we used typical augmentations like random cropping and horizontal flipping. To balance representation, a random sample of images was drawn for each class before training.

Model Architectures. We used three convolutional backbones—ResNet-50, MobileNetV2, and VGG-19—adapted to the number of output classes in RadImageNet. Pretrained weights were used where available. Training was performed with mini-batches of 64 using stochastic gradient descent with momentum and weight decay. The learning rate started at 0.001 and decayed in steps at predefined epochs. Models were trained for a fixed number of epochs, with validation accuracy monitored to prevent overfitting.

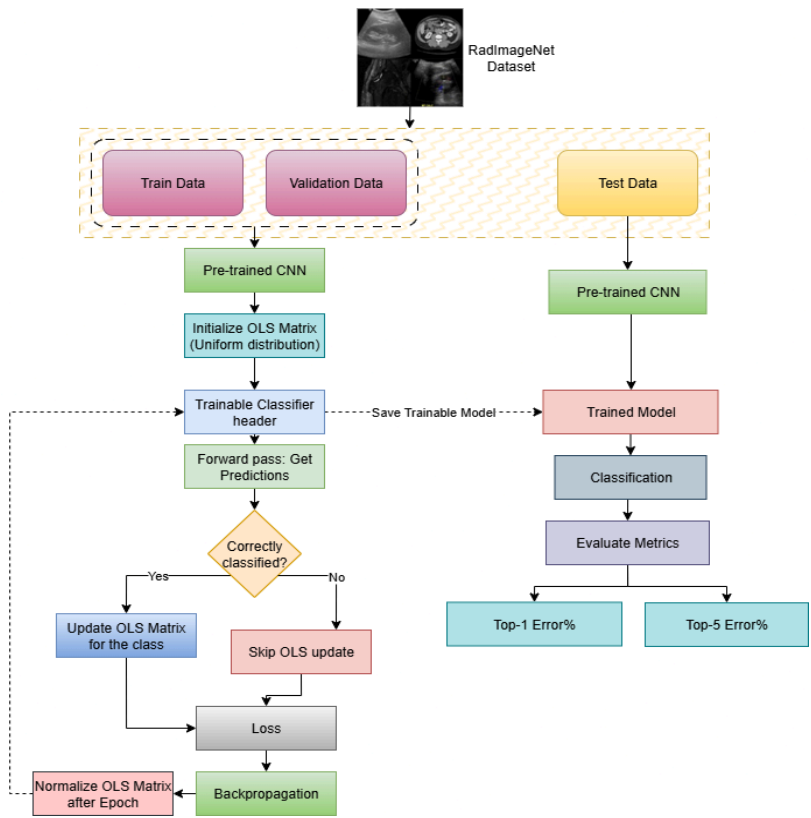

**Fig.1.** Overview of the training and evaluation pipeline using Online Label Smoothing (OLS).

Online Label Smoothing Implementation. Unlike static label smoothing, which applies the same uniform distribution to all non-target classes, OLS updates class-wise soft labels based on model predictions during training as shown in Fig.1. At each epoch, when a sample of class i was correctly classified, its predicted probability vector was added to an accumulator for that class. After processing all samples, the accumulated vectors were normalized to yield a probability distribution representing the confusions of the current model for class i. These evolving soft labels were stored in a matrix and used in the next epoch as additional supervision.

Evaluation Metrics. Model outputs were assessed on the RadImageNet test set using Top-1 and Top-5 accuracy, as well as calibration metrics including Expected Calibration Error (ECE), and average confidence.

## 4    Experimental Work

Following the approach proposed by [1], we adopt the online label smoothing mechanism where the soft labels are dynamically updated throughout training. Instead of relying on a static smoothed target distribution, we maintain a class-level soft label that evolves over time. During each training epoch, if a sample $x_i$ is correctly classified by the model, the predicted probability distribution $p(x_i)$ is used to update the soft label corresponding to the true class $y_i$. These updated soft labels are then employed in subsequent epochs to supervise the model training.

Let $T$ denote the total number of training epochs. Following the formulation in [1], we maintain a set of soft label matrices $S = \{S^0, S^1, ...., S^t, .... S^{T-1}\}$, where each matrix $S^t \in R^{K \times K}$ encodes class-level soft labels at epoch t. Each column in $S^t$ represents the soft label distribution for a particular class. During the $t_{th}$ epoch of training, for a given sample ($x_i, y_i$), the soft label from the previous epoch $S^{t-1}_{y_i}$ (i.e., the column corresponding to class $y_i$) is used to construct the target distribution for supervision. . The training loss of the model can be represented by

$$L_{soft} = - \sum_{k=1}^{K} S^{t-1}_{y_i,k} \cdot log p(k|x_i) \qquad (1)$$

The hard label distribution $q$ for each sample is defined such that $q(k \mid x_i) = 1$ if $k = y_i$, and $q(k \mid x_i) = 0$ otherwise. Using this, the standard cross-entropy loss for hard label supervision is expressed as:

$$L_{hard} = - \sum_{k=1}^{K} q(k|x_i) log p(k|x_i) \qquad (2)$$

While it is theoretically possible to train the model solely using the soft labels updated during training, doing so from random initialization often leads to poor convergence due to the absence of explicit hard targets in early stages. To address this, we adopt a hybrid loss formulation that incorporates both hard and soft labels. As proposed in [1], the final training objective is defined as a weighted combination of the two:

$$L = \alpha \cdot L_{hard} + (1 - \alpha) \cdot L_{soft} \quad (3)$$

where $\alpha \in [0, 1]$ is a balancing hyperparameter that controls the trade-off between the hard and soft supervision signals.

We update the class-wise soft labels at each training epoch t using the model's own predictions. These updated soft labels are then used to supervise the model in the subsequent epoch t+1. Specifically, the soft label matrix $S^t \in R^{K \times K}$ is initialized as a zero matrix at the start of each epoch. For every training sample $(x_i, y_i)$ that is correctly classified, we use its predicted probability vector $p(x_i)$ to update the column of St corresponding to class $y_i$. The update is defined as:

$$S^t_{y_i,k} = S^t_{y_i,k} + p(k|x_i), \forall k \in \{1,...,K\}. \quad (4)$$

After processing all correctly predicted samples, the matrix $S^t$ is normalized column-wise to form valid probability distributions:

$$S^t_{y_i,k} \leftarrow \frac{S^t_{j,k}}{\sum_{l=1}^{K} S^t_{j,l}} \quad (5)$$

The resulting normalized soft label matrix $S^t$ is then used to supervise the model in epoch t+1. As there are no prior predictions available during the first epoch (t=0), the initial soft label matrix $S^0$ is set to a uniform distribution across all classes, as done in [1].

## 5     Results

Table 1 summarizes the Top-1 and Top-5 classification errors on RadImageNet for ResNet-50, MobileNetV2, and VGG-19 under four training regimes: hard (one-hot) labels, standard label smoothing (LS), Teacher-free Knowledge Distillation (Tf-KD), and Online Label Smoothing (OLS). Across all three architectures, OLS consistently achieves the lowest error rates. The gains in Top-1 accuracy are moderate but consistent: for example, OLS reduces the Top-1 error of ResNet-50 by roughly 1–2% (absolute) compared to hard labels. Both LS and Tf-KD provide modest improvements over the hard-label baseline, but OLS yields the largest improvements. In medical applications even minor improvements in accuracy can have significant consequences — for instance, reducing misclassifications in high-risk conditions like tumors or pulmonary disease has a direct impact on patient outcomes. In our experiments, the most significant advantage of OLS appears in the Top-5 metric. For each backbone, the Top-5 error under OLS is much lower than under LS or Tf-KD.

Table 1. Various Model Performance across different architecture

| Backbones | Hard Label | | LS | | Tf-KD | | OLS | |
|---|---|---|---|---|---|---|---|---|
| | Top-1 Err(%) | Top-5 Err(%) | Top-1 Err(%) | Top-5 Err(%) | Top-1 Err(%) | Top-5 Err(%) | Top-1 Err(%) | Top-5 Err(%) |
| ResNet-50 | 24.27 | 5.20 | 23.84 | 6.60 | 25.17 | 5.90 | 23.80 | 4.61 |
| MobileNet V2 | 25.01 | 5.04 | 21.79 | 6.71 | 26.49 | 6.49 | 24.43 | 4.34 |
| VGG-19 | 30.38 | 5.84 | 29.26 | 5.62 | 30.24 | 5.73 | 27.19 | 5.37 |

Also, we quantified calibration performance using Expected Calibration Error (ECE) and mean prediction confidence. Results for ResNet-50 on three training schemes: baseline (hard labels), standard label smoothing (LS), and Online Label Smoothing (OLS) are presented in Table 2. The best ECE (0.0151) was obtained by the OLS method with a more conservative mean confidence (0.7906), which exhibited significantly better agreement of predicted probabilities with actual correctness in comparison to baseline or LS. The findings indicate that OLS not only improves classification accuracy but also provides well-calibrated confidence estimates, a factor of particular significance in medical image applications.

Table 2. Calibration performance on ResNet-50

| Method | ResNet-50 | |
|---|---|---|
| | ECE | Average Confidence |
| Baseline | 0.1537 | 0.8783 |
| LS | 0.0611 | 0.8303 |
| OLS | 0.0151 | 0.7906 |

Fig. 2. compares the penultimate-layer embeddings from the baseline (hard labels), LS, and OLS models (the same pattern applies to all backbones). The OLS features clusters much more tightly, more compact cluster groups for each class, with good separation between different classes. By contrast, the no-smoothing baseline model has scattered clusters with significant overlap, and the LS model falls in between.

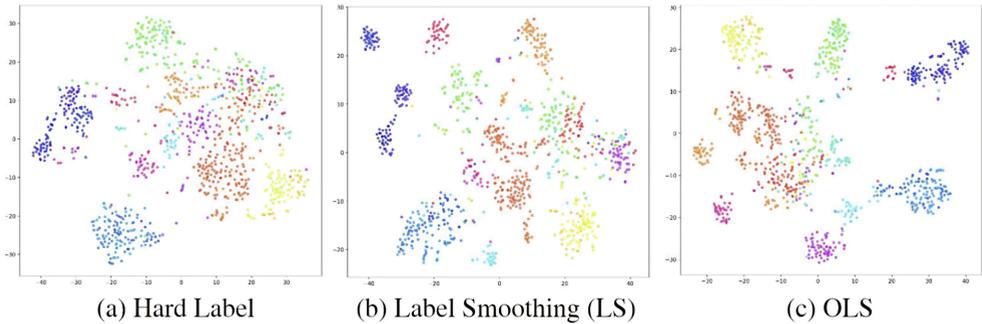

(a) Hard Label      (b) Label Smoothing (LS)      (c) OLS

**Fig. 2.** Visualization of the layer representations of ResNet-56 on RadImageNet training set using t-SNE

Overall, the experiments demonstrate that OLS acts as an effective regularizer for medical image classification. By adaptively generating soft targets, OLS prevents over-confidence and overfitting. The approach continually enhances Top-1 and Top-5 accuracy on RadImageNet on different architectures, surpassing hard-label training, uniform LS, and Tf-KD. The particularly significant reduction in Top-5 error indicates that OLS helps the model achieve more nuanced differences between the highest-ranked predictions.

## 6 Conclusion

Online Label Smoothing, dynamically changed label distributions with respect to real-time performance of the model while training, hence solved the limitations inherent in traditional label smoothing. Evident particularly in much smaller Top-5 error rates than under normal static label smoothing and hard-label training methods, tests on the comprehensive RadImageNet dataset using a range of CNN architectures—ResNet-50, MobileNetV2, and VGG-19—repeatedly demonstrated improved classification results. In clinical scenarios where tiny distinctions among similar diseases are common, OLS makes for improved secondary class predictions, something that is extremely significant.

Aside from improved classification, our experiments also illustrate that OLS is more calibrated than baseline and LS models. One of the most important reasons that this improvement can be traced back to is the way the two methods handle non-target classes. Classical LS distributes the same probability mass equally over all the incorrect classes, without consideration of the fact that some classes are visually or semantically closer to the target ground truth than others. By contrast, OLS learns soft labels dynamically from the model's own mistakes at training time, assigning more weight to likely confusions and very little to irrelevant classes. The adaptive distribution generates more calibrated supervision, minimizing Expected Calibration Error (ECE) and regulating average confidence, resulting in predictions that more accurately indicate correctness. This process enables OLS to identify clinically significant uncertainty instead of overestimating certainty, a characteristic particularly valuable in medical imaging when diagnostic credibility is critical.

From a clinical standpoint, more accurate calibration with OLS has a number of applications. Calibrated confidence estimates enable clinicians to treat model outputs as accurate markers of correctness, promoting improved diagnostic decision-making. Practically, such models would be applied to prioritize high-risk patients and avoid overconfident misclassification and resulting unnecessary procedures.

Future research may apply this framework to multi-center data sets and examine hybrid strategies that integrate OLS with other calibration techniques to further align model predictions with clinical requirements.

computer vision to medical imaging. arXiv preprint arXiv:2206.08833 (2022). https://doi.org/10.48550/arXiv.2206.08833.